\documentclass[sigconf]{acmart}

\AtBeginDocument{%
  }

\setcopyright{acmcopyright}
\copyrightyear{2022}
\acmYear{2022}
\setcopyright{acmlicensed}\acmConference[HCMA '22]{Proceedings of the 3rd International Workshop on Human-Centric Multimedia Analysis}{October 10--14, 2022}{Lisboa, Portugal}
\acmBooktitle{Proceedings of the 3rd International Workshop on Human-Centric Multimedia Analysis (HCMA '22), October 10--14, 2022, Lisboa, Portugal}
\acmPrice{15.00}
\acmDOI{10.1145/3552458.3556447}
\acmISBN{978-1-4503-9492-5/22/10}

\usepackage{multirow}
\begin{document}

\title{Dual Domain-Adversarial Learning for Audio-Visual Saliency Prediction}

%
\author{Yingzi Fan}
\affiliation{%
  \institution{Xidian University}
  \city{Xi'an}
  \country{China}
}
\email{15091172525@163.com}

\author{{Longfei Han}$^\ast$}
\affiliation{%
  \institution{Beijing Technology and Business University}
  \city{Beijing}
  \country{China}
}
\email{draflyhan@gmail.com}
\thanks{$^\ast$ Corresponding author}

\author{Yue Zhang}
\affiliation{%
 \institution{Xi’an Jiaotong University}
 \city{Xi’an}
 \country{China}
}
\email{yuezhang@xjtu.edu.cn}

\author{Lechao Cheng}
\affiliation{%
  \institution{Zhejiang Lab}
  \city{Hangzhou}
  \country{China}
  }
 \email{chenglc@zhejianglab.com}

\author{Chen Xia}
\affiliation{%
  \institution{Northwestern Polytechnical University}
  \city{Xi’an}
  \country{China}
}
\email{cxia@nwpu.edu.cn}

\author{Di Hu}
\affiliation{%
  \institution{Renmin University of China}
  \city{Beijing}
  \country{China}
}
\email{hdui831@mail.nwpu.edu.cn}



\begin{abstract}
Both visual and auditory information are valuable to determine the salient regions in videos. 
Deep convolution neural networks (CNN) showcase strong capacity in coping with the audio-visual saliency prediction task. Due to various factors such as shooting scenes and weather, there often exists moderate distribution discrepancy between source training data and target testing data. The domain discrepancy induces to performance degradation on target testing data for CNN models.
This paper makes an early attempt to tackle the unsupervised domain adaptation problem for audio-visual saliency prediction. We propose a dual domain-adversarial learning algorithm to mitigate the domain discrepancy between source and target data. First, a specific domain discrimination branch is built up for aligning the auditory feature distributions. Then, those auditory features are fused into the visual features through a cross-modal self-attention module. The other domain discrimination branch is devised to reduce the domain discrepancy of visual features and audio-visual correlations implied by the fused audio-visual features. Experiments on public benchmarks demonstrate that our method can relieve the performance degradation caused by domain discrepancy. 
\end{abstract}




\maketitle

\section{Introduction}
Visual saliency prediction (VSP) aims to extract important regions from input images, which is inspired by the selective attention mechanism of humans. It has a very wide range of application in computer vision tasks,  such as video summarization \cite{marat}, stream compression \cite{2013saliency}, video monitoring \cite{2010predictive}, etc.
In real world, visual signals are usually accompanied by audio signals, which can also provide valuable clues for locating salient regions.
Recently, the audio-visual saliency prediction (AVSP) task has attracted extensive research interest. Benefiting from their strong representation capacity, deep convolutional neural networks (CNN) bring huge development to this field~\cite{tavakoli2019dave,tsiami2020stavis}.
However, CNN models trained with source data usually do not perform well on target data due to the data distribution shift.
This paper focuses on studying the unsupervised domain adaptation problem for AVSP.

Existing AVSP methods~\cite{tavakoli2019dave,tsiami2020stavis,yuan2021bio} mainly concentrate on modeling the feature representations of auditory and visual inputs and the interactions between the two kinds of inputs. They all employ a two-branch framework to encode the audio and video inputs. A 3D convolution branch is leveraged to extract visual features from video inputs. \cite{tavakoli2019dave} utilizes a 3D convolution branch to extract auditory feature representations from the log mel-spectrogram frames of the audio inputs, while \cite{tsiami2020stavis,yuan2021bio,jain2020avinet} implement the audio representation modeling with SoundNet~\cite{aytar2016soundnet}. Fusing the visual and auditory features is the other critical point in AVSP. \cite{tsiami2020stavis} simply uses the concatenation operation to fuse the two kinds of features. \cite{yuan2021bio} devises attention mechanisms to explore the spatial and temporal dependencies between the two features.
These methods achieve excellent performance under the circumstance that testing data obeys the same distribution with training data. However, the unsupervised adaptation of them to unseen testing data is still under research.


\begin{figure*}[t]
\center
\includegraphics[width=0.7\textwidth]{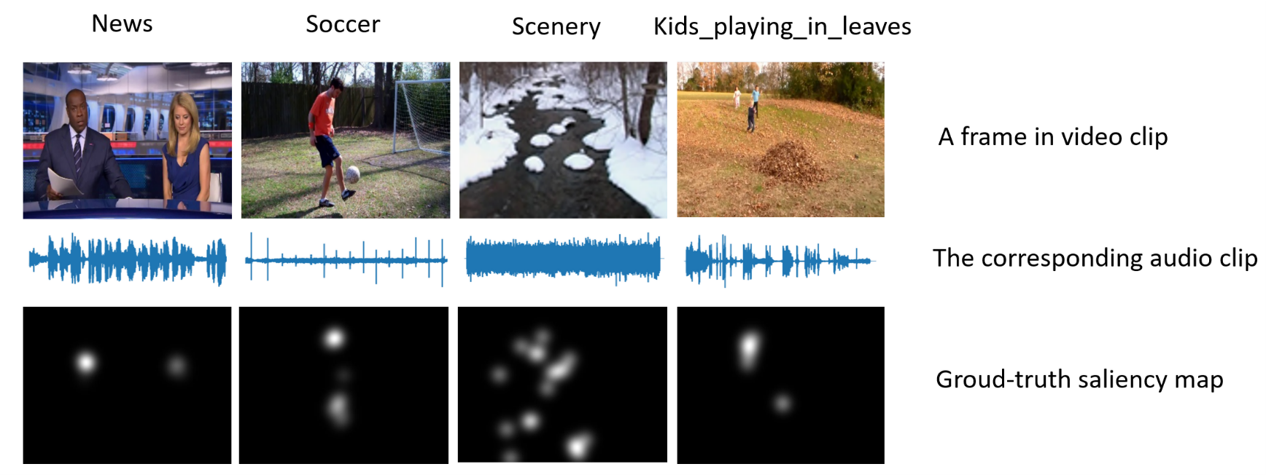}
\caption{Examples for visualizing the variance of audio and video data.} \label{fig1}
\end{figure*} 

The exploration of visual contrast information and audio-visual interaction is critical to settle the AVSP task. However, these factors can be easily influenced by the shooting conditions. Objects having relatively higher contrast to background stuffs are more prone to attract attention of humans. The identities of such objects are flexible and vary as the change of shooting scenes. For example, in a speaking video, the mouth region usually draws the highest attention, while the attractive regions are completely different in other social activities such as sport and dance scenes. 
Visual data differences induced by such factors can degrade the performance of AVSP methods learned with stationary source data. 
It is likely to contain crucial cues for recognizing salient regions that the audio information accompanying video data. However, the intrinsic attributes of sound data such as pitch, loudness, and timbre are diversified among videos. 
On the other hand, the relevance between audio and salient regions is variant according to different types of videos. For example, the sound resource locations are usually consistent with salient regions in speaking videos as indicated by the `news' example of Fig.~\ref{fig1}; salient regions can not be thoroughly identified out according to sounds in more general social videos like the 
`soccer' example of Fig.~\ref{fig1}; in natural scene videos (`scenery' and `kids playing in leaves' of Fig.~\ref{fig1}), the audio data is generated by multiple sound resources, and the connection between audio and salient regions is sometimes ambiguous. Due to the above factors, the performance of AVSP methods can be hampered by the domain discrepancy caused by variances of the audio data and the interaction between the audio and the video data.

\begin{figure*}[t]
\center
\includegraphics[width=0.95\textwidth]{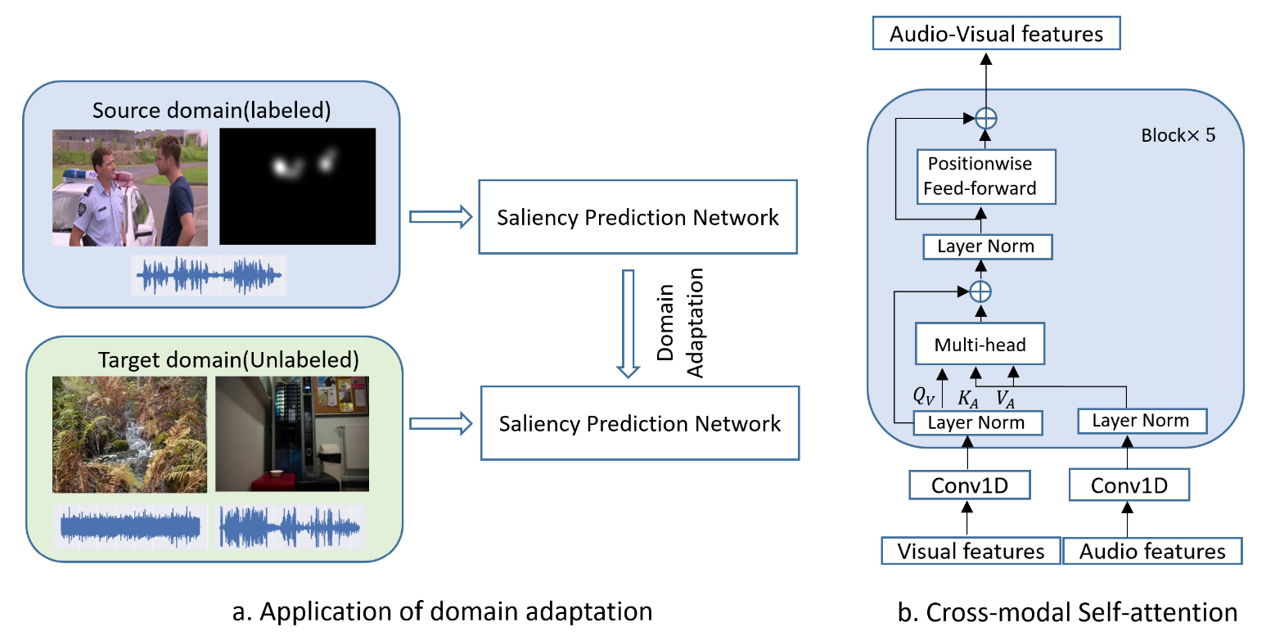}
\caption{A brief introduction to our proposed method.} \label{fig2}
\end{figure*}

To deal with the issue of domain discrepancy, we propose a dual domain-adversarial learning framework on the basis of the unsupervised domain adaptation paradigm (Fig.~\ref{fig2} (a)). Following existing methods~\cite{tsiami2020stavis,yuan2021bio}, we set up independent branches for extracting features from audio and video inputs respectively.
First, an auditory feature domain discriminator is set up to eliminate the impacts of audio data variance. It guides the optimization of the audio feature extraction branch via adversarial training and helps to produce domain-invariant audio features. 
Then, a cross-modal self-attention module as shown in Fig.~\ref{fig2} (b) is devised to fuse auditory features into visual features extracted by a 3D convolution backbone. The other domain discriminator is adopted to learn fused audio-visual features having a uniform distribution across source and target domains. This is helpful for mitigating the domain discrepancy caused by the variance of visual features and the relation between auditory and visual features.
To synthesize the effect of domain discrepancy, we set up two experimental settings: 1) videos captured under different types of scenes are regarded as samples of source and target domains; 2) datasets collected from different sites are regarded as source and target domains.
Extensive experiments on two settings demonstrate that our proposed method is effective in making up the performance reduction caused by domain discrepancy. 

In summary, the main contributions of this paper are as follows.
\begin{itemize}
    \item [1)] We make an early attempt to tackle the unsupervised domain adaptation problem for audio-visual saliency prediction.
    \item [2)] A dual domain adversarial learning framework is proposed for aligning the audio-visual feature distributions between source and target domains.
    \item [3)] We conduct extensive experiments on cross-scene and cross-dataset adaptation settings, validating the efficacy of our proposed method.
\end{itemize}


\section{Related Work}
\subsection{Video Saliency}
With the development of deep convolutional neural networks, a vast number of saliency prediction methods are proposed, and they achieve excellent performance on the image saliency prediction \cite{huang2015salicon,zhang2015exploiting,pan2017salgan,wang2017deep} and video saliency prediction \cite{bak2017spatio,jiang2018deepvs,wang2018revisiting,min2019tased} task.
In video saliency prediction, many recent methods utilize an LSTM to extract spatial and temporal visual features. Wang et al. \cite{wang2018revisiting} propose a CNN-LSTM network architecture with an attention mechanism. Linardos et al. \cite{linardos2019simple} add a ConvLSTM to an existing neural network and wrap a convolutional layer by using a temporal exponential moving average(EMA) operation. Some existing models have explored 3D-CNN architecture to learn visual features. TASED-Net \cite{min2019tased} is a 3D fully-convolutional network, including an encoding network to extract spatiotemporal features and a prediction network to produce a single saliency map. STSConvNet \cite{bak2017spatio} extract the 
temporal and spatial information by using a two-stream network architecture, then fusing them to get a saliency map. Fang et al. \cite{fang2022densely} proposed a densely nested top-down flows (DNTDF)-based framework to predict pixel-wise salient object regions. \cite{zhuge2022salient} designed an ICON to integrate the features from the micro level and macro level. In \cite{cong2021rrnet}, a relational reasoning network is proposed for SOD in optical RSIs. \cite{cong2018review} review some visual saliency detection algorithms with comprehensive information. Liu et al. \cite{liu2020learning} incorporated multi-modal information to learn attention by integrating self-attention and each other’s attention.  \cite{zhang2020uc} is the first work to employ uncertainty for RGB-D saliency detection, which generates multiple saliency maps by modeling human annotation uncertainty. A survey of RGB-D salient object detection is provided by Zhou et al.\cite{zhou2021rgb}.



\subsection{Audio-Video Saliency}

Some early audio-visual saliency prediction methods have been explored for application-specific\cite{ruesch2008multimodal,schauerte2011multimodal,chen2014audio,liao2015audeosynth}. They employ traditional signal processing techniques for visual saliency and audio localization. Coutrot et al. \cite{coutrot2014audiovisual} extract static and dynamic video features by using Gabor filters, then weighing the face regions appropriately and combining the visual saliency map to generate the final audio-visual saliency map. Also, Min et al. \cite{min2016fixation} propose an audio-visual model to predict the salient region in the scenes containing moving and sound-generating objects. They generate the saliency map by fusing the audio, spatial, and temporal attention maps. 

However, only a few works have explored the end-to-end network for audio-visual saliency fixation prediction based on deep learning. Tavakoli et al. \cite{tavakoli2019dave} adopt two 3D-ResNet as the backbone of the two modalities respectively, and their outputs are catenated for encoding. Tsiami et al. \cite{tsiami2020stavis} propose STAViS, which employs a multi-modal network that combines visual and auditory information at multiple stages. Furthermore, they investigate the fusion approaches such as computing the
cosine similarity, taking the weighted inner product, and bilinear transformation to fuse the audio modality. Jain et al. \cite{jain2020avinet} propose a hierarchical structure for audio-video saliency prediction. They analyze three fusion mechanisms, including simple concatenation, bilinear fusion, and transformer-based fusion, to fuse video and audio cues. Zhu et al. \cite{zhu2021lavs} propose a lightweight audio-visual
saliency model for the audio-visual saliency fixation prediction task. Although these methods show excellent performance, they do not involve the study of unsupervised domain adaptation when the distribution of unlabeled testing data differs from training data.

\subsection{Domain Adaptation}
Unsupervised domain adaptation(UDA) is targeted at aligning the data distribution of labeled source domain and unlabeled target domain. Thus, the model trained in the source domain can be directly migrated to the target domain without significant performance degradation. 
The core issue in UDA is how to align the distributions of source and target data. A large number of UDA algorithms have been proposed for classical computer vision tasks such as image classification and semantic segmentation, based on image translation~\cite{hoffman2018cycada,murez2018image}, maximum mean discrepancy minimization~\cite{wang2020rethink}, or domain adversarial learning~\cite{ganin2015unsupervised,liu2021adversarial}. 
In this paper, we concentrate on the unsupervised domain adaptation for audio-visual saliency prediction which has not been studied in previous literature. 

\begin{figure*}[t]
\center
\includegraphics[width=1\textwidth]{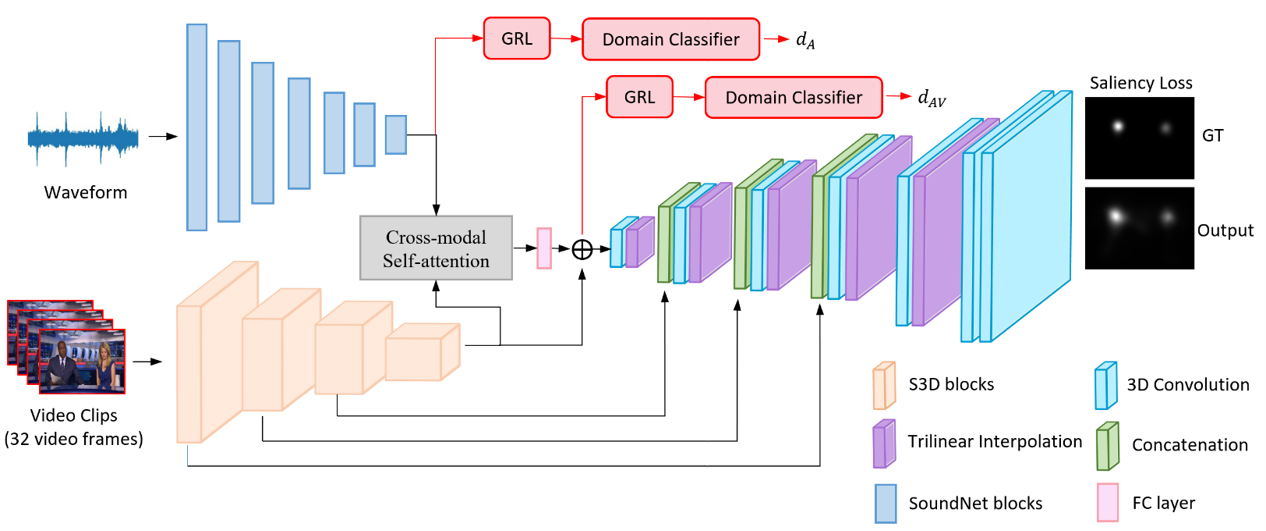}
\caption{Overall architecture of the proposed model.} \label{fig3}
\end{figure*}

\section{Method}
This paper aims to tackle the unsupervised domain adaptation for AVSP. Assume the source domain data be $\{(\mathbf A^s_i, \mathbf V^s_i)\}_{i=1}^{N_s}$, where $\mathbf A^s_i$ and $\mathbf V^s_i$ denotes the video and audio of the $i$-th sample respectively. 
$N_s$ represents the number of source samples.
The ground-truth saliency map for the $i$-th sample in the source dataset is denoted as $\mathbf G^s_i$. 
For the target data, only a subset of unlabeled samples  $\{(\mathbf A^t_j, \mathbf V^t_j)\}_{j=1}^{N_t}$. $N_t$ represents the number of target samples.
The goal of this paper is to learn a CNN model capable of achieving high AVSP performance on the target domain data. A dual domain adversarial training pipeline is proposed for addressing the above problem. Technical details are introduced in following subsections. 

\subsection{Architecture Overview}
Fig. 3 illustrates the proposed architecture for cross-modal audio-visual saliency prediction. The proposed architecture consists of a visual branch to extract visual features, an audio branch to extract audio features, a cross-modal self-attention module to obtain correlated meaningful audio-visual features for saliency prediction, and a decoder module to decode the audio-visual features and get the final saliency map. Domain adaptation modules are added to make the model learn generalizable features.
We input consecutive video frames $\mathbf{V}$ and the corresponding sampled audio sequences $\mathbf{A}$ into the visual and audio branches, respectively, and the final output is a saliency prediction map $\mathbf{P}$. Each domain classifier outputs a binary result to indicate whether the input video comes from the target domain or the source domain.

For visual branch, we use S3D \cite{xie2018rethinking} as the backbone to extract visual features. 
The size of input video frames $\mathbf{V}$ is $C_0\times F _0 \times H _0 \times W_0$, 
where $F_0 = 32 $ is the number of frames and $C_0 = 3$ is the number of frame channels. $H_0$ and $W_0$ represents the height and width of each frame, respectively. The output of the visual branch is $\mathbf{O}_{\rm V}$. The size of $\mathbf{O}_{\rm V}$ is $C\times F \times H \times W$, where $C = 1024$, $F=\frac{F _0}{8}$, $H=\frac{H _0}{32}$, and $W=\frac{W _0}{32}$. 

For the audio branch, we adopt SoundNet~\cite{aytar2016soundnet} to encode the representation of the audio sequence, which is a 1-D fully convolutional network and is initially proposed for audio classification. 
The size of pre-processed audio sequence $\mathbf{A}$ is $ 1\times T _a \times 1$ , and the size of the output features $\mathbf{O}_{\rm A}$ is $1024 \times 3 \times 1$.

For audio-visual fusion, inspired by \cite{tsai2019multimodal}, we use the cross-modal self-attention mechanism to fuse two modalities. $\mathbf{O}_{\rm A}$ and $\mathbf{O}_{\rm V}$ through dimensional changes are input to the cross-modal self-attention fusion module. For the details, see Section 3.2. Then the fusion information goes through a fully-connected layer and is added with $\mathbf{X}_{\rm V}$ to obtain the salient audio-visual features $\mathbf{O}_{\rm AV}$.

Afterwards, $\mathbf{O}_{\rm AV}$ are fed into a decoder consisting of six decoding layers. The first decoding layer contains a 3D convolution and a trilinear interpolation layer. Each of the next three decoding layers contains a concatenation, a 3D convolutional, and a trilinear upsampling layer. The concatenation operation combines the output of the visual block and the output of the last decoder layer along the temporal dimension. The fourth decoding layer is similar to the first layer. The last decoding layer is composed of two 3D convolutional layers, which project the channel and temporal dimension to 1. Finally, the single map goes through a sigmoid
activation function to obtain a saliency map $\mathbf{P}$.

Domain adaptation modules are applied to make the knowledge acquired in the labeled source domain to be better generalized in the unlabeled target domain. Specifically, the branch consisting of a gradient reversal layer and a domain classifier is added after the audio output and the output of the fusion module, respectively.

In the following, we mainly describe the cross-modal self-attention mechanism of the fusion module(see Section 3.2) and domain adaptation(see Section 3.3).

\subsection{Cross-modal Self-attention}
For the fusion of two modalities, previous methods such as computing the cosine similarity between two modalities, simple concatenation, bilinear fusion and so on, can not fully explore the correlation between two modalities. In order to obtain meaningful audio-visual correlation information, we adopt a cross-modal self-attention mechanism, which enables visual modality to receive information from audio modality selectively. The source audio modality is used to reconstruct the information of visual modality and the cross-modal interaction between vision and audio is realized. Details about the cross-modal self-attention are as follows.

The output features of the visual branch $\mathbf{O}_{\rm V}$ are first processed with a max pooling layer and are flattened to $\mathbf{x}_{\rm V} \in {\mathbf{R} ^{C \times (HW) }}$. Then the $\mathbf{x}_{\rm V}$ passes through a 1-D convolution and then are collapsed as a vector $ \mathbf{X}_{\rm V} \in {\mathbf{R} ^{T_V \times d_V }} $, and the audio features $\mathbf{O}_{\rm A} $ are passed through a 1-D convolution and then are collapsed as a vector  $ \mathbf{X}_{\rm A} \in {\mathbf{R} ^{T_A \times d_A }} $. Where $T_V = H \times W = 42$, $T_A = 3$, and $d_V = d_A = 512$. $ T_{(\cdot)} $ is the length of the sequence and $ d_{(\cdot)} $ is the feature dimension. The cross-modal attention from $\mathbf{X}_{\rm A} $ to $\mathbf{X}_{\rm V} $ is defined as $\mathbf{Y}_{\rm{A \rightarrow V}}$. The calculation process is as below.

\begin{equation}
    \mathbf{Y}_{ \rm{A} \rightarrow \rm{V}} = softmax (\frac{\mathbf{Q}_{\rm{V}}\mathbf{K}_{\rm{A}}^{\rm{T}}}{\sqrt{d_k}})\mathbf{V}_{\rm{A}}
\end{equation}


where $\mathbf{Q} _{\rm V} = {\mathbf{X}_{\rm V}}{\mathbf{W}_{\rm{Q_V}}}, \mathbf{\mathbf{K} _{\rm A} = {\mathbf{X}_{\rm A}}{\mathbf{W}_{\rm{K_A}}}}$ and  $\mathbf{V} _{\rm A} = {\mathbf{X}_{\rm A}}{\mathbf{W}_{\rm{V_A}}}$ represent query, key and value in the cross-modal self-attention module. $\mathbf{W}_{\rm {Q_V}} \in {\mathbf{R}^{{d_V} \times {d_k}}}, \mathbf{W}_{\rm{K_A}} \in {\mathbf{R}^{{d_A} \times {d_k}}}$ and $ \mathbf{W}_{\rm{V_A}} \in {\mathbf{R}^{{d_A} \times {d_v}}}$ are weights. 

In this way, we get the correlation matrix between the audio sequence and the corresponding video, which plays a great role in determining the saliency region of the video. Then, $\mathbf{Y}_{\rm{A \rightarrow V}}$ is added back into $\mathbf{X}_{\rm V}$, and a residual multi-layer perceptron is attached for further feature enhancement.
The complete calculation process is illustrated in Fig.~\ref{fig2} (b). 

\subsection{Domain Adaptation}

Our unsupervised domain adaptation module is constituted by a gradient inversion layer and a domain classifier. The gradient inversion layer is located between the feature extraction module and the domain classifier. 
During the error backpropagation process, this layer reverses the gradient direction, namely making the feature extraction module confuse the domain classifier. 
Such a manner avoids the two-stage training process of adversarial training. 
For the domain classifier, three $1 \times 1$ spatial convolution layers are adopted to compress the feature dimension. Afterwards, three fully-connected layers are attached, producing the domain classification score.

We setup two separate domain classifiers for audio features and fused audio-visual features respectively. The domain classifier of audio features is denoted as $D_{\rm{A}}$, and that of audio-visual features is denoted as $D_{\rm{AV}}$.
Suppose the audio feature  and the audio-visual feature extracted from a source domain sample ($\mathbf A^s$, $\mathbf V^s$) be $\mathbf O^s_{\rm{A}}$ and $\mathbf O^s_{\rm{AV}}$, respectively.
$\mathbf O^t_{\rm{A}}$ and $\mathbf O^t_{\rm{AV}}$ denotes the audio feature  and the audio-visual feature of a target domain sample  ($\mathbf A^t$, $\mathbf V^t$), respectively.
$\mathbf O^s_{\rm{A}}$ and $\mathbf O^t_{\rm{A}}$ are fed into the audio domain classifier $D_{\rm{A}}$, resulting in prediction scores $d^{s}_{\rm{A}}$ and $d^{t}_{\rm{A}}$, respectively.
On the other hand, $D_{\rm{AV}}$ predicts domain classification scores $d^{s}_{\rm{AV}}$ and $d^{t}_{\rm{AV}}$ from $\mathbf O^s_{\rm{AV}}$ and $\mathbf O^t_{\rm{AV}}$, respectively. 
The domain classification loss for $D_{\rm{A}}$ is defined as $L_{\rm{A}}$ and the loss for $D_{\rm{AV}}$ is $L_{\rm{AV}}$ : 

\begin{eqnarray}
\label{eq:adv-audio} L_{\rm{A}}( d^s_{\rm{A}}, d^t_{\rm{A}}) &= -\log(d^s_{\rm{A}} ) -\log(1-d^t_{\rm{A}}), \\
\label{eq:adv-av} L_{\rm{AV}}( d^s_{\rm{AV}}, d^t_{\rm{AV}}) &= -\log(d^s_{\rm{AV}}) -\log(1-d^t_{\rm{AV}}).
\end{eqnarray}

\subsection{Loss Function}
The source domain images are employed for training the saliency prediction model with the Kullback-Leibler (KL) divergence. The training loss $L_s$ on source domain images is as below:
\begin{equation}
\label{eq:sup} L_s(\mathbf P^s,\mathbf G^s) = \sum_{i=1}^{|\mathbf P^s|}\mathbf G^s(i)\log(\epsilon + \frac{\mathbf G^s(i)}{\mathbf P^s(i) + \epsilon}),
\end{equation}
where $\mathbf{P}^s$ is the saliency map predicted from $\mathbf A^s$ and $\mathbf V^s$,  $\mathbf{G}$ is the corresponding ground-truth map, and $\epsilon$ is a regularization parameter. $\mathbf G^s(i)$ and $\mathbf P^s(i)$ represents the $i$-th pixel of $\mathbf G^s$ and $\mathbf P^s$, respectively. $|\mathbf P^s|$ denotes the number of pixels in $\mathbf P^s$.  
The final loss function is formed by summing up (\ref{eq:adv-audio}), (\ref{eq:adv-av}), and (\ref{eq:sup}):
\begin{equation}
L = L_s + L_{\rm{A}} + L_{\rm{AV}}.
\end{equation}

\section{Experiments}
In this section, we validate the effectiveness of our method on multiple benchmarks. For evaluating the performance of unsupervised domain adaptation, we synthesize different data domains according to shooting scenarios or dataset sources.

\subsection{Datasets and Evaluation Metrics}
We carry out the experiments on six audio-visual eye-tracking datasets: DIEM \cite{mital2011clustering}, AVAD \cite{min2016fixation}, Coutrot1 \cite{coutrot2014saliency}, Coutrot2 \cite{coutrot2016multimodal}, SumMe \cite{gygli2014creating}, and ETMD \cite{koutras2015perceptually}. DIEM contains 84 video clips, including commercials, documentaries, game trailers, movie trailers, etc. Coutrot1 contains 60 dynamic nature scene clips such as moving objects, landscapes and faces. Coutrot2 contains 15 clips of 4 people in the meeting. AVAD contains 45 short clips with audio-visual scenes such as dancing, guitar playing, birds singing, etc. ETMD contains 12 videos from six different Hollywood films. SumMe contains 25 unstructured videos. These eye-tracking data are all recorded via the Eyelink eye-tracker from viewers.
\begin{table}
\caption{Number of videos in each dataset.}\label{tab1}
\center
\begin{tabular}{c|c|c}
\hline
Dataset &  Train & Valid./Test\\
\hline
Faces &  30 & 14\\
Nature &  18 & 10\\
Social & 58 & 15\\
\hline
\end{tabular}
\end{table}
For the evaluation of the domain adaptation, we divide these datasets into two settings based on their shooting scenes. One of the settings is `DIEM, AVAD and Coutrot1', which consists of parts of AVAD, Coutrot1 and DIEM datasets. The three datasets contain multiple scenarios. The other is `Faces, Nature and Social'. The three datasets are constructed by selecting data from the existing six datasets according to the content of video frames. `Faces' contains 30 clips with faces, and the corresponding audio is human voice. `Nature' contains 18 clips with landscapes and scenery. `Social' contains 58 clips, including dancing, guitar playing, playing football, etc. The number of videos in each dataset is summarized in Table~\ref{tab1}.

\begin{figure*}[t]
\center
\includegraphics[width=0.7\textwidth]{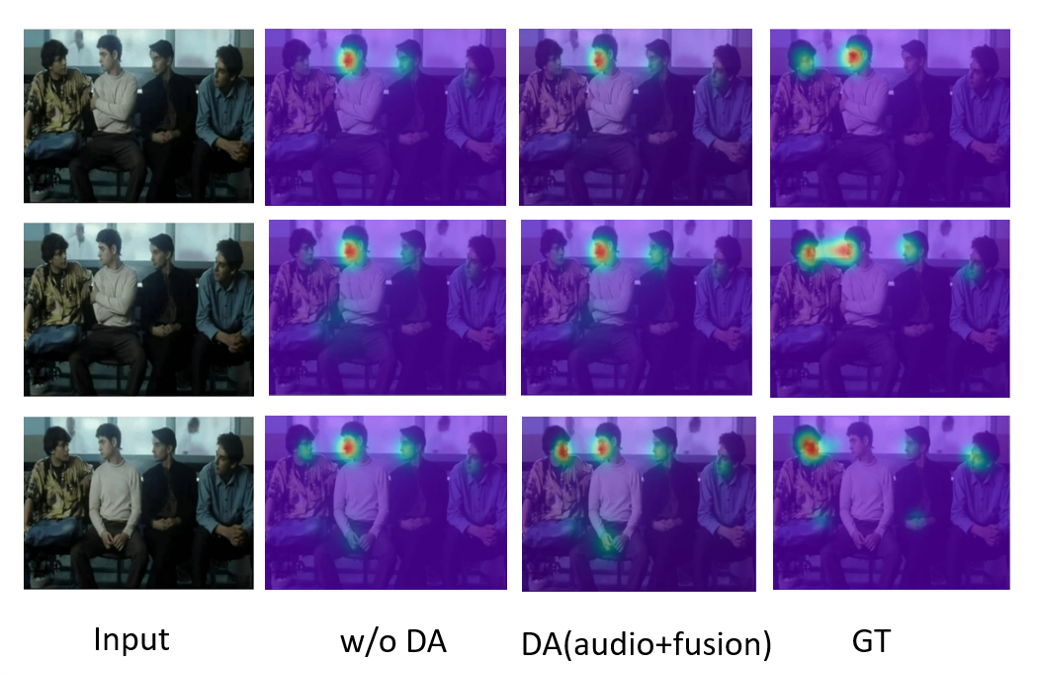}
\caption{Sample frames from Coutrot1 dataset, and the corresponding ground truth, and saliency maps without adding DA, and adding DA after the audio branch and fusion module.} \label{fig5}
\end{figure*}

\begin{table*}[t]
\centering
\caption{Performance of our proposed method under various domain adaptation settings.}\label{tab:da}
\begin{tabular}{c | c c c c c c } 
\hline
\multicolumn{7}{c}{Target data: Nature}\\
\hline
Source data & Approach & CC &  NSS &  sAUC & AUC-J & SIM \\
\hline
\multirow{3}{*}{Faces}&  w/o DA &0.4099 &  1.670 & 0.5981 &  0.8442 &  0.3587 \\
& DA(audio) & 0.4352  &  1.785 &  \textbf{0.6021} & 0.8591  & 0.3574  \\
&DA(audio + fusion)       &  \textbf{0.4488} &  \textbf{1.843} &  0.5988  & \textbf{0.8625}  &  \textbf{0.3608} \\
\hline
\multirow{3}{*}{Social}&  w/o DA & 0.477 & 1.9711  & 0.6019  &  0.8717 &  0.3742 \\
& DA(audio) & 0.4951  & 2.064  &  0.6016 &  0.8721 &  \textbf{0.3964} \\
&DA(audio + fusion)  &  \textbf{0.4985} &  \textbf{2.084} & \textbf{ 0.6083} &  \textbf{0.8727} &  0.3833 \\
\hline
\multicolumn{7}{c}{} \\
\hline
\multicolumn{7}{c}{Source data: AVAD} \\
\hline
Target data & Approach & CC &  NSS &  sAUC & AUC-J & SIM \\
\hline
\multirow{3}{*}{Coutrot1} &  w/o DA &  0.4685 &  2.274 & 0.5881  &  0.8673 & 0.3962 \\
 & DA(audio)  & 0.4749  &  2.298 & 0.5926  & 0.8716  & 0.3962 \\
& DA(audio + fusion)       &  \textbf{0.4891} & \textbf{2.364}  &  \textbf{0.5927} &  \textbf{ 0.8723} & \textbf{0.4057} \\
\hline
\multirow{3}{*}{DIEM}&  w/o DA &  0.5640 &  2.281 & 0.6744  &  0.8857 & 0.4710 \\
& DA(audio) & 0.5692 &  2.300 &  0.6782 &  0.8898 & 0.4712 \\
& DA(audio + fusion)       & \textbf{0.5758} &  \textbf{2.315} &  \textbf{0.6797} &  \textbf{0.8911} &  \textbf{0.4714} \\
\hline
\end{tabular}
\end{table*}

We use five evaluation metrics \cite{bylinskii2018different}, including Linear Correlation Coefficient (CC), AUC-Judd (AUC-J), Similarity Metric (SIM), shuffled AUC (s-AUC), and Normalized Scanpath Saliency (NSS) to evaluate the performance of our model.

\subsection{Experimental Setup}
In our experiment, for the visual branch, we use the weights pre-trained on DHF1K \cite{wang2018revisiting} to initialize the model. For the audio feature extraction network, we initialize the weights from the SoundNet pre-trained in the sound localization task \cite{aytar2016soundnet}. The learning rate is set to be $10^{-4}$, and the batch size is 8 when training the model without domain adaptation. When adding the domain adaptation, the learning rate is set to be $10^{-5}$, and the batch size is set to 6. In all experiments, we adopt the Adam optimizer to update network parameters. 


\subsection{Results}
\begin{figure*}[t]
\center
\includegraphics[width=1\textwidth]{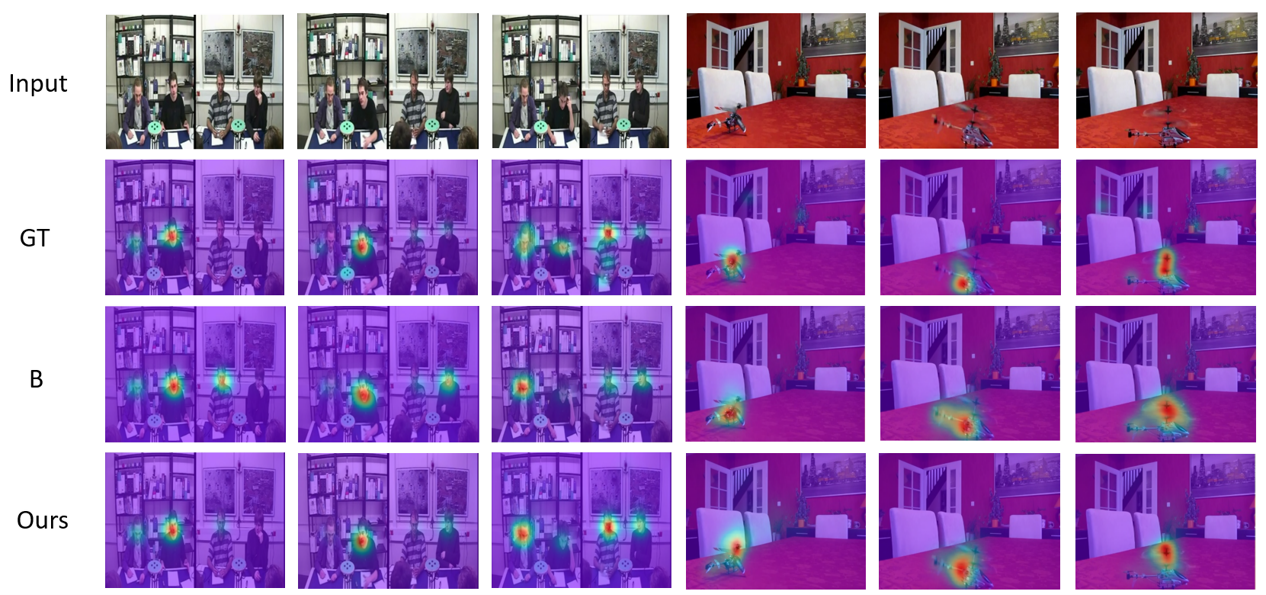}
\caption{Sample frames from Coutrot2 and Coutrot1 datasets, and the corresponding ground truth, and saliency maps generated by our cross-model self-attention fusion and bilinear fusion for comparisons.} \label{fig4}
\end{figure*}

\begin{table*}[t]
\caption{ Experimental results of using different auditory and visual feature fusion methods. Evaluation metrics are calculated by averaging results on ETMD, SumMe, Coutrot1, Coutrot2, DIEM and AVAD.}\label{tab:fusion}
\center
\begin{tabular}{c|ccccc}
\hline
Methods & CC &  NSS &  sAUC & AUC-J & SIM \\
\hline
C &     \textbf{0.6167} & 3.367 & \textbf{0.6993} & 0.9155 & 0.4422 \\
B &      0.6090& 3.412 & 0.6965 & 0.9152 & 0.4427  \\
CM & 0.6138 & \textbf{3.428} & 0.6757 & \textbf{0.9191} & \textbf{0.4735} \\
\hline
\end{tabular}
\end{table*}

\begin{table*}[t]
\caption{ Comparision results. Evaluation metrics are calculated by averaging results on ETMD, SumMe, Coutrot1, Coutrot2, DIEM and AVAD.}\label{tab:comparision}
\center
\begin{tabular}{c|ccccc}
\hline
{} & CC &  NSS &  sAUC & AUC-J & SIM \\
\hline
STAViS &     0.5644 & 2.9683 & 0.6584 & 0.854 & 0.4345 \\
AViNet &      0.6118& 3.3683 & \textbf{0.6998} & 0.9178 & 0.4445  \\
Ours & \textbf{0.6138} & \textbf{3.428} & 0.6757 & \textbf{0.9191} & \textbf{0.4735} \\
\hline
\end{tabular}
\end{table*}
We conduct domain adaptation experiments on several different combinations of datasets. Specifically, we take Faces/Social datasets as the source domain, and Nature dataset as the target domain, forming domain adaptation settings: Faces$\rightarrow$Nature and Social$\rightarrow$Nature. The result of not adding the domain adaptation(DA) is tested, denoted by w/o DA, which means the model is trained on the source domain and is tested directly on the target domain. DA(audio) and DA(audio + fusion) represent adding DA after the audio branch and adding DA after the audio branch and fusion module. The results are shown in Table2. Among  DIEM, Coutrot1 and AVAD datasets, we take AVAD dataset as the source domain, and DIEM/Coutrot1 datasets as the target domain, forming another two domain adaptation settings: AVAD$\rightarrow$DIEM and AVAD$\rightarrow$Coutrot1. The experimental results are shown in Table~\ref{tab:da}.

The experimental results show that the performance of the model on the unlabeled target domain datasets is improved by adding DA module. This indicates that the model with DA has learned the domain-invariant characteristics, and the discrepancy between source and target domain is mitigated. We also visualize the results in Fig.\ref{fig5}, where the model with DA module performs better.

\subsection{Ablation Study}
\noindent \textbf{Efficacy of Dual Domain-Adversarial Learning.} We analyze the improvement of the model’s performance by adding DA (domain adaptation) in different parts of the model so as to determine the most appropriate position for adding DA. We perform ablation experiments in the following two cases: (1) adding DA after the audio branch (DA(audio)); (2) adding DA after the audio branch and fusion module (DA(audio + fusion)). Table~\ref{tab:da} shows the test results of the model in the above two cases. The results show that adding DA after the audio branch and fusion module performs better.
\vspace{2mm}

\noindent \textbf{Efficacy of Cross-Modal Self-Attention.} We verify the effectiveness of our proposed fusion strategy on six audio-visual datasets by comparing it with existing fusion methods. One is concatenating visual and auditory features, and the other is bilinear fusion. Table~\ref{tab:fusion} shows the averaged evaluation metrics. `C' represents simple concatenation, `B' represents bilinear fusion, and `CM' represents our proposed fusion strategy based on cross-modal self-attention mechanism. It can be seen from Table~\ref{tab:fusion} that the proposed fusion method is superior to other methods on NSS, AUC-J, and SIM, demonstrating the effectiveness of our proposed fusion method. Furthermore, we compare the fusion methods qualitatively. In Fig.\ref{fig4}, the first row depicts the sample frames from Coutrot2 and Coutrot1. The ground-truth saliency maps are presented in the second row. The final two rows include the saliency maps generated by bilinear fusion and our cross-model self-attention fusion. As shown in 
the Figure 5, our results are closer to the ground truth. What's more, to prove our model's efficacy without DA for AVSP task, we compare our model with other AVSP methods. Table~\ref{tab:comparision} shows the averaged evaluation metrics on six audio-visual datasets. As is shown, our method obtains better performance on several metrics.

\section{Conclusion}
In this work, we propose a dual domain-adversarial learning algorithm to solve the problem of unsupervised domain adaptation in audio-visual saliency detection. 
Specifically, we establish two domain discrimination branches to align the distribution of auditory features and to reduce the domain differences in visual features and audio-visual correlation. 
Experimental results show that our method can relieve the performance degradation caused by domain discrepancy on the audio-visual saliency prediction task. 

\begin{acks}
This work is supported in part by the National Major Scientific Instruments and Equipments Development Project of National Natural Science Foundation of China No. 62027813 and in part by the Key Program of the National Natural Science Foundation of China under Grant No. 62036005. And it is partially supported by the National Natural Science Foundation of China (Grant No. 62106235), by the Exploratory Research Project of Zhejiang Lab(2022PG0AN01), by the Zhejiang Provincial Natural Science Foundation of China (LQ21F020003).
\end{acks}
\bibliographystyle{unsrt}
\bibliography{paper}

\begin{thebibliography}{10}

\bibitem{marat}
Sophie Marat, Mick{\"a}el Guironnet, and Denis Pellerin.
\newblock Video summarization using a visual attention model.
\newblock In {\em 2007 15th European Signal Processing Conference}, pages
  1784--1788. IEEE, 2007.

\bibitem{2013saliency}
Hadi Hadizadeh and Ivan~V Baji{\'c}.
\newblock Saliency-aware video compression.
\newblock {\em IEEE Transactions on Image Processing}, 23(1):19--33, 2013.

\bibitem{2010predictive}
Fahad Fazal~Elahi Guraya, Faouzi~Alaya Cheikh, Alain Tremeau, Yubing Tong, and
  Hubert Konik.
\newblock Predictive saliency maps for surveillance videos.
\newblock In {\em 2010 Ninth International Symposium on Distributed Computing
  and Applications to Business, Engineering and Science}, pages 508--513. IEEE,
  2010.

\bibitem{tavakoli2019dave}
Hamed~R Tavakoli, Ali Borji, Esa Rahtu, and Juho Kannala.
\newblock Dave: A deep audio-visual embedding for dynamic saliency prediction.
\newblock {\em arXiv preprint arXiv:1905.10693}, 2019.

\bibitem{tsiami2020stavis}
Antigoni Tsiami, Petros Koutras, and Petros Maragos.
\newblock Stavis: Spatio-temporal audiovisual saliency network.
\newblock In {\em Proceedings of the IEEE/CVF Conference on Computer Vision and
  Pattern Recognition}, pages 4766--4776, 2020.

\bibitem{yuan2021bio}
Yuan Yuan, Hailong Ning, and Bin Zhao.
\newblock Bio-inspired audio-visual cues integration for visual attention
  prediction.
\newblock {\em arXiv preprint arXiv:2109.08371}, 2021.

\bibitem{jain2020avinet}
Samyak Jain, Pradeep Yarlagadda, Ramanathan Subramanian, and Vineet Gandhi.
\newblock Avinet: Diving deep into audio-visual saliency prediction.
\newblock {\em arXiv e-prints}, pages arXiv--2012, 2020.

\bibitem{aytar2016soundnet}
Yusuf Aytar, Carl Vondrick, and Antonio Torralba.
\newblock Soundnet: Learning sound representations from unlabeled video.
\newblock {\em Advances in neural information processing systems}, 29, 2016.

\bibitem{huang2015salicon}
Xun Huang, Chengyao Shen, Xavier Boix, and Qi~Zhao.
\newblock Salicon: Reducing the semantic gap in saliency prediction by adapting
  deep neural networks.
\newblock In {\em Proceedings of the IEEE international conference on computer
  vision}, pages 262--270, 2015.

\bibitem{zhang2015exploiting}
Jianming Zhang and Stan Sclaroff.
\newblock Exploiting surroundedness for saliency detection: a boolean map
  approach.
\newblock {\em IEEE transactions on pattern analysis and machine intelligence},
  38(5):889--902, 2015.

\bibitem{pan2017salgan}
Junting Pan, Cristian~Canton Ferrer, Kevin McGuinness, Noel~E O'Connor, Jordi
  Torres, Elisa Sayrol, and Xavier Giro-i Nieto.
\newblock Salgan: Visual saliency prediction with generative adversarial
  networks.
\newblock {\em arXiv preprint arXiv:1701.01081}, 2017.

\bibitem{wang2017deep}
Wenguan Wang and Jianbing Shen.
\newblock Deep visual attention prediction.
\newblock {\em IEEE Transactions on Image Processing}, 27(5):2368--2378, 2017.

\bibitem{bak2017spatio}
Cagdas Bak, Aysun Kocak, Erkut Erdem, and Aykut Erdem.
\newblock Spatio-temporal saliency networks for dynamic saliency prediction.
\newblock {\em IEEE Transactions on Multimedia}, 20(7):1688--1698, 2017.

\bibitem{jiang2018deepvs}
Lai Jiang, Mai Xu, Tie Liu, Minglang Qiao, and Zulin Wang.
\newblock Deepvs: A deep learning based video saliency prediction approach.
\newblock In {\em Proceedings of the european conference on computer vision
  (eccv)}, pages 602--617, 2018.

\bibitem{wang2018revisiting}
Wenguan Wang, Jianbing Shen, Fang Guo, Ming-Ming Cheng, and Ali Borji.
\newblock Revisiting video saliency: A large-scale benchmark and a new model.
\newblock In {\em Proceedings of the IEEE Conference on Computer Vision and
  Pattern Recognition}, pages 4894--4903, 2018.

\bibitem{min2019tased}
Kyle Min and Jason~J Corso.
\newblock Tased-net: Temporally-aggregating spatial encoder-decoder network for
  video saliency detection.
\newblock In {\em Proceedings of the IEEE/CVF International Conference on
  Computer Vision}, pages 2394--2403, 2019.

\bibitem{linardos2019simple}
Panagiotis Linardos, Eva Mohedano, Juan~Jose Nieto, Noel~E O'Connor, Xavier
  Giro-i Nieto, and Kevin McGuinness.
\newblock Simple vs complex temporal recurrences for video saliency prediction.
\newblock {\em arXiv preprint arXiv:1907.01869}, 2019.

\bibitem{fang2022densely}
Chaowei Fang, Haibin Tian, Dingwen Zhang, Qiang Zhang, Jungong Han, and Junwei
  Han.
\newblock Densely nested top-down flows for salient object detection.
\newblock {\em Science China Information Sciences}, 65(8):1--14, 2022.

\bibitem{zhuge2022salient}
Mingchen Zhuge, Deng-Ping Fan, Nian Liu, Dingwen Zhang, Dong Xu, and Ling Shao.
\newblock Salient object detection via integrity learning.
\newblock {\em IEEE Transactions on Pattern Analysis and Machine Intelligence},
  2022.

\bibitem{cong2021rrnet}
Runmin Cong, Yumo Zhang, Leyuan Fang, Jun Li, Yao Zhao, and Sam Kwong.
\newblock Rrnet: Relational reasoning network with parallel multiscale
  attention for salient object detection in optical remote sensing images.
\newblock {\em IEEE Transactions on Geoscience and Remote Sensing}, 60:1--11,
  2021.

\bibitem{cong2018review}
Runmin Cong, Jianjun Lei, Huazhu Fu, Ming-Ming Cheng, Weisi Lin, and Qingming
  Huang.
\newblock Review of visual saliency detection with comprehensive information.
\newblock {\em IEEE Transactions on circuits and Systems for Video Technology},
  29(10):2941--2959, 2018.

\bibitem{liu2020learning}
Nian Liu, Ni~Zhang, and Junwei Han.
\newblock Learning selective self-mutual attention for rgb-d saliency
  detection.
\newblock In {\em Proceedings of the IEEE/CVF conference on computer vision and
  pattern recognition}, pages 13756--13765, 2020.

\bibitem{zhang2020uc}
Jing Zhang, Deng-Ping Fan, Yuchao Dai, Saeed Anwar, Fatemeh~Sadat Saleh, Tong
  Zhang, and Nick Barnes.
\newblock Uc-net: Uncertainty inspired rgb-d saliency detection via conditional
  variational autoencoders.
\newblock In {\em Proceedings of the IEEE/CVF conference on computer vision and
  pattern recognition}, pages 8582--8591, 2020.

\bibitem{zhou2021rgb}
Tao Zhou, Deng-Ping Fan, Ming-Ming Cheng, Jianbing Shen, and Ling Shao.
\newblock Rgb-d salient object detection: A survey.
\newblock {\em Computational Visual Media}, 7(1):37--69, 2021.

\bibitem{ruesch2008multimodal}
Jonas Ruesch, Manuel Lopes, Alexandre Bernardino, Jonas Hornstein, Jos{\'e}
  Santos-Victor, and Rolf Pfeifer.
\newblock Multimodal saliency-based bottom-up attention a framework for the
  humanoid robot icub.
\newblock In {\em 2008 IEEE International Conference on Robotics and
  Automation}, pages 962--967. IEEE, 2008.

\bibitem{schauerte2011multimodal}
Boris Schauerte, Benjamin K{\"u}hn, Kristian Kroschel, and Rainer Stiefelhagen.
\newblock Multimodal saliency-based attention for object-based scene analysis.
\newblock In {\em 2011 IEEE/RSJ International Conference on Intelligent Robots
  and Systems}, pages 1173--1179. IEEE, 2011.

\bibitem{chen2014audio}
Yanxiang Chen, Tam~V Nguyen, Mohan Kankanhalli, Jun Yuan, Shuicheng Yan, and
  Meng Wang.
\newblock Audio matters in visual attention.
\newblock {\em IEEE Transactions on Circuits and Systems for Video Technology},
  24(11):1992--2003, 2014.

\bibitem{liao2015audeosynth}
Zicheng Liao, Yizhou Yu, Bingchen Gong, and Lechao Cheng.
\newblock Audeosynth: music-driven video montage.
\newblock {\em ACM Transactions on Graphics (TOG)}, 34(4):1--10, 2015.

\bibitem{coutrot2014audiovisual}
Antoine Coutrot and Nathalie Guyader.
\newblock An audiovisual attention model for natural conversation scenes.
\newblock In {\em 2014 IEEE International Conference on Image Processing
  (ICIP)}, pages 1100--1104. IEEE, 2014.

\bibitem{min2016fixation}
Xiongkuo Min, Guangtao Zhai, Ke~Gu, and Xiaokang Yang.
\newblock Fixation prediction through multimodal analysis.
\newblock {\em ACM Transactions on Multimedia Computing, Communications, and
  Applications (TOMM)}, 13(1):1--23, 2016.

\bibitem{zhu2021lavs}
Dandan Zhu, Defang Zhao, Xiongkuo Min, Tian Han, Qiangqiang Zhou, Shaobo Yu,
  Yongqing Chen, Guangtao Zhai, and Xiaokang Yang.
\newblock Lavs: A lightweight audio-visual saliency prediction model.
\newblock In {\em 2021 IEEE International Conference on Multimedia and Expo
  (ICME)}, pages 1--6. IEEE, 2021.

\bibitem{hoffman2018cycada}
Judy Hoffman, Eric Tzeng, Taesung Park, Jun-Yan Zhu, Phillip Isola, Kate
  Saenko, Alexei Efros, and Trevor Darrell.
\newblock Cycada: Cycle-consistent adversarial domain adaptation.
\newblock In {\em International conference on machine learning}, pages
  1989--1998. PMLR, 2018.

\bibitem{murez2018image}
Zak Murez, Soheil Kolouri, David Kriegman, Ravi Ramamoorthi, and Kyungnam Kim.
\newblock Image to image translation for domain adaptation.
\newblock In {\em Proceedings of the IEEE Conference on Computer Vision and
  Pattern Recognition}, pages 4500--4509, 2018.

\bibitem{wang2020rethink}
Wei Wang, Haojie Li, Zhengming Ding, and Zhihui Wang.
\newblock Rethink maximum mean discrepancy for domain adaptation.
\newblock {\em arXiv preprint arXiv:2007.00689}, 2020.

\bibitem{ganin2015unsupervised}
Yaroslav Ganin and Victor Lempitsky.
\newblock Unsupervised domain adaptation by backpropagation.
\newblock In {\em International conference on machine learning}, pages
  1180--1189. PMLR, 2015.

\bibitem{liu2021adversarial}
Xiaofeng Liu, Zhenhua Guo, Site Li, Fangxu Xing, Jane You, C-C~Jay Kuo, Georges
  El~Fakhri, and Jonghye Woo.
\newblock Adversarial unsupervised domain adaptation with conditional and label
  shift: Infer, align and iterate.
\newblock In {\em Proceedings of the IEEE/CVF International Conference on
  Computer Vision}, pages 10367--10376, 2021.

\bibitem{xie2018rethinking}
Saining Xie, Chen Sun, Jonathan Huang, Zhuowen Tu, and Kevin Murphy.
\newblock Rethinking spatiotemporal feature learning: Speed-accuracy trade-offs
  in video classification.
\newblock In {\em Proceedings of the European conference on computer vision
  (ECCV)}, pages 305--321, 2018.

\bibitem{tsai2019multimodal}
Yao-Hung~Hubert Tsai, Shaojie Bai, Paul~Pu Liang, J~Zico Kolter, Louis-Philippe
  Morency, and Ruslan Salakhutdinov.
\newblock Multimodal transformer for unaligned multimodal language sequences.
\newblock In {\em Proceedings of the conference. Association for Computational
  Linguistics. Meeting}, volume 2019, page 6558. NIH Public Access, 2019.

\bibitem{mital2011clustering}
Parag~K Mital, Tim~J Smith, Robin~L Hill, and John~M Henderson.
\newblock Clustering of gaze during dynamic scene viewing is predicted by
  motion.
\newblock {\em Cognitive computation}, 3(1):5--24, 2011.

\bibitem{coutrot2014saliency}
Antoine Coutrot and Nathalie Guyader.
\newblock How saliency, faces, and sound influence gaze in dynamic social
  scenes.
\newblock {\em Journal of vision}, 14(8):5--5, 2014.

\bibitem{coutrot2016multimodal}
Antoine Coutrot and Nathalie Guyader.
\newblock Multimodal saliency models for videos.
\newblock In {\em From Human Attention to Computational Attention}, pages
  291--304. Springer, 2016.

\bibitem{gygli2014creating}
Michael Gygli, Helmut Grabner, Hayko Riemenschneider, and Luc~Van Gool.
\newblock Creating summaries from user videos.
\newblock In {\em European conference on computer vision}, pages 505--520.
  Springer, 2014.

\bibitem{koutras2015perceptually}
Petros Koutras and Petros Maragos.
\newblock A perceptually based spatio-temporal computational framework for
  visual saliency estimation.
\newblock {\em Signal Processing: Image Communication}, 38:15--31, 2015.

\bibitem{bylinskii2018different}
Zoya Bylinskii, Tilke Judd, Aude Oliva, Antonio Torralba, and Fr{\'e}do Durand.
\newblock What do different evaluation metrics tell us about saliency models?
\newblock {\em IEEE transactions on pattern analysis and machine intelligence},
  41(3):740--757, 2018.

\end{thebibliography}
\end{document}